\documentclass{article}

\usepackage{graphicx} 
\usepackage[main,preprint]{neurips_2026}
\usepackage[utf8]{inputenc} 
\usepackage[T1]{fontenc}    
\usepackage{hyperref}       
\usepackage{url}            
\usepackage{booktabs}       
\usepackage{amsfonts}       
\usepackage{nicefrac}       
\usepackage{microtype}      
\usepackage{subcaption}     
\usepackage{xcolor}         
\usepackage{amsmath}
\usepackage{enumitem}
\usepackage{float}
\usepackage{longtable}
\hypersetup{hidelinks}
\title{EvoMem: Memory-Augmented Evolution for Code Optimization}
\author{%
Viktor Volkov $^{1}$ \\ 
\And
Valentin Khrulkov $^{1}$ \\
\And
Andrey V. Galichin $^{1}$ \\
\And
Danil Sivtsov $^{1}$ \\
\And
Nikita Glazkov $^{1}$ \\
\And
Olga Volkova $^{1}$ \\
\And 
Konstantin Pchelin $^{1}$ \\
\And
Iaroslav Bespalov $^{1}$ \\
\And
Dmitry V. Dylov $^{1,3}$ \\
\And
Petr Anokhin $^{1,2}$ \\
\And
Ivan Oseledets $^{1,2}$ \\[0.5em]
$^1$AXXX, Moscow, Russia \\
$^2$Lomonosov Moscow State University, Moscow, Russia \\
$^3$Applied AI Institute, Moscow, Russia \\
}

\date{April 2026}

\begin{document}

\maketitle

\begin{abstract}
Successful mutation strategies in evolutionary code search may contain reusable knowledge that is useful beyond a single run, and in some cases may transfer across related tasks and domains. However, existing LLM-driven evolutionary frameworks largely discard such knowledge, repeatedly rediscovering similar ideas and limiting opportunities for cross-run and cross-task learning. We introduce EvoMem, a persistent memory architecture for LLM-based evolutionary program search that captures and reuses candidate mutation knowledge. EvoMem converts successful mutation events into structured, task-aware advice for future runs. It operates in two phases: after each run, it extracts and stores promising ideas with provenance, and during subsequent evolution, it retrieves a small set of relevant instructions based on the current task and program context to guide mutation. Across geometric optimization, multi-hop question answering, GPU kernel optimization, and related benchmarks, our experiments show positive average improvements in target metrics or search speed for most evaluated settings, while also revealing variability across tasks. Overall, EvoMem provides evidence that persistent memory can reduce some redundant exploration and improve the reuse and adaptation of successful strategies in LLM-driven evolutionary search.
\end{abstract} 

\section{Introduction}
LLM-based evolutionary code search has become a promising way to automate
program improvement. Systems such as AlphaEvolve combine LLM-generated code
mutations with evolutionary selection: they retrieve candidate programs,
prompt an LLM to propose a modification, evaluate the result, and retain
successful variants in a population \citep{novikov2025alphaevolvecodingagentscientific}.
This evaluator-in-the-loop structure is especially attractive for research and
optimization tasks in which proposed ideas can be expressed as code and tested
automatically. AlphaEvolve has already demonstrated this pattern on matrix
multiplication, mathematical search, and systems optimization, showing that
LLMs can be useful mutation operators when they are embedded in a disciplined
search loop.

However, current evolutionary code-search systems mostly treat each run as a
self-contained search process. Successful runs produce more than a final
program: they also reveal tactics such as useful decompositions, numerical
stabilization tricks, pruning rules, data-layout changes, or prompting
strategies that may remain valuable beyond the original task. In standard
evolutionary pipelines, this knowledge is usually left implicit in the evolved
programs and logs. When the system starts a new run, even on a related problem,
it must often rediscover similar ideas through broad search.

Naively reusing prior artifacts is not enough. A concrete program is usually
tied to a particular task, metric, and interface, while simply mixing many
experiments into one population can make retrieval less precise and can expose
the mutation model to irrelevant context. What is needed is a mechanism that
abstracts useful mutation knowledge from past runs, preserves enough provenance
to keep it auditable, and retrieves only a bounded amount of relevant advice
during future mutations.

We present EvoMem, a persistent memory architecture for LLM-based evolutionary
program search. EvoMem is implemented as an extension to GigaEvo, an
open-source implementation of an AlphaEvolve-style optimization pipeline
\citep{khrulkov2025gigaevoopensourceoptimization}. After an evolutionary run,
EvoMem extracts successful mutation events into structured memory records,
deduplicates and enriches them, and stores them with task context and
provenance. During later evolution, it retrieves a small set of relevant memory
instructions based on the current task and program context and inserts them
into the mutation prompt as advice rather than as a hard constraint.

This paper makes the following contributions:
\begin{itemize}[leftmargin=*]
    \item We introduce EvoMem, a persistent memory mechanism that stores and
    reuses mutation knowledge across LLM-based evolutionary code-search runs.
    \item We describe a two-phase memory pipeline: an offline write phase that
    extracts, filters, deduplicates, and stores successful ideas with
    provenance, and an online read phase that retrieves bounded task-relevant
    advice during mutation.
    \item We implement EvoMem on top of GigaEvo and evaluate it on transfer
    settings spanning geometric optimization, multi-hop question answering, GPU kernel optimization, and scientific-code
    tuning.
    \item We provide empirical evidence that memory reuse can accelerate some
    later evolutionary runs and can support cross-problem transfer, with
    improvements appearing as higher target metrics, faster
    search, or both depending on the task.
\end{itemize}

\begin{figure}[t]
    \centering
    \includegraphics[width=\textwidth]{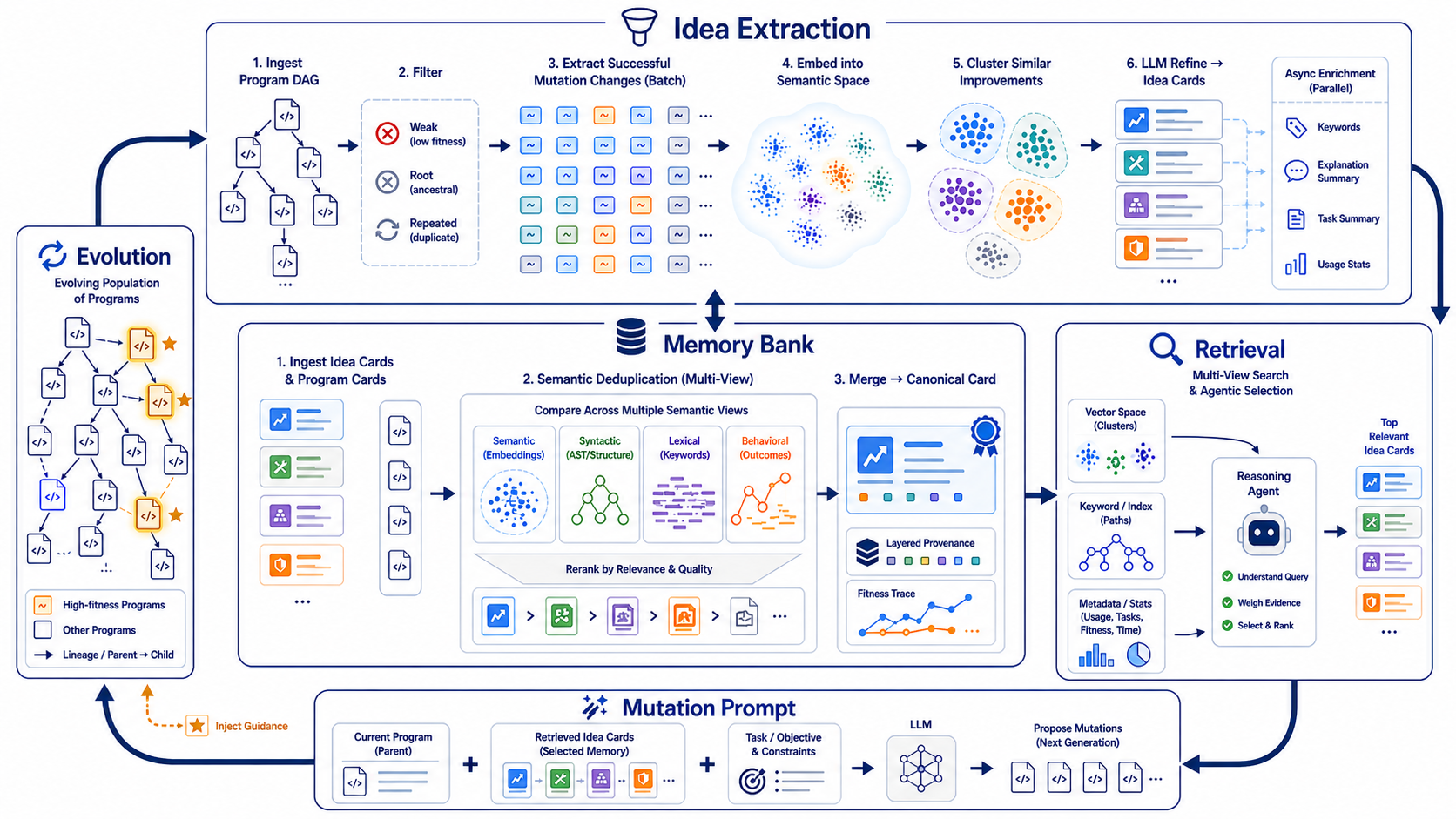}
    \caption{EvoMem pipeline. Successful mutation events from completed runs are distilled into structured memory cards with provenance, stored in a persistent memory layer, and retrieved as bounded advice for future mutations without changing the underlying evolutionary loop.}
    \label{fig:title-image}
\end{figure}

\section{Related Work}

Recent AI-driven scientific-discovery systems combine code generation, retrieval, external tools, and experimental feedback to automate parts of the research workflow \citep{eger2025transforming,ren2025scientificintelligence,hasib2025processcentric}. Within this broader landscape, the line of work most relevant to this paper is evolutionary program and code search, where systems such as AlphaEvolve and CodeEvolve iteratively modify candidate programs, evaluate them, and retain successful variants \citep{novikov2025alphaevolvecodingagentscientific,assumpcao2025codeevolve}. Our focus is narrower than the general scientific-discovery literature: we study how reusable knowledge can be extracted from prior evolutionary runs and reintroduced as persistent memory in later mutations.

Further research on AlphaEvolve capabilities has shown that this approach can rediscover and improve strong existing solutions across multiple domains, including mathematical reasoning, combinatorial search, and systems optimization \citep{Georgiev2025MathematicalEAG,nagda2026reinforcedgenerationcombinatorialstructures,Li2026DiscoveringMLI,Vitvitskyi2026MiningGAJ,Kramar2026BuildingPPB}. Recent variants also show evolutionary approaches being adapted to specialized domains such as realistic CAD generation and quantum-circuit T-count optimization, while ImprovEvolve changes the evolved program interface around proposing, improving, and perturbing candidate solutions \citep{elistratov2026cadevolvecreatingrealistic,fisher2026llmguidedevolutionarysearchalgebraic,Kravatskiy2026ImprovEvolveAAE}. Follow-up work on open or alternative implementations suggests that the search loop itself can be improved: smaller RL-trained models may outperform larger closed ensembles in some settings, task decomposition can make evolution more efficient and interpretable, and adaptive metrics, sampling, or meta-evolution can further improve performance \citep{Wang2025ThetaEvolveTLA,li2025fm,Liu2026EvoXMFH}.

The closest neighboring systems are those that combine evolutionary code search with domain knowledge or explicit memory. AlphaEvolve and related systems such as ShinkaEvolve and MAPPS show that iterative program modification and evaluation can support scientific and algorithmic discovery in specialized settings \citep{novikov2025alphaevolvecodingagentscientific,shinkaevolve2025,mapps2025}. Nearby examples include GeoEvolve, which combines evolutionary search with domain knowledge, and AgenticSciML, which adds retrieval-augmented method memory to scientific machine-learning workflows \citep{luo2025geoevolve,jiang2025agenticsciml}. Most other domain-specific agent systems focus on chemistry, biology, or materials pipelines rather than on persistent cross-run memory for evolutionary code search \citep{xiao2024cellagent,ghafarollahi2024protagents,lin2022esm}.

Agentic memory architectures such as A-MEM and General Agentic Memory likewise retrieve and update agent experience at task time \citep{xu2025amemagenticmemoryllm,yan2025generalagenticmemorydeep}. EvoMem differs by storing provenance-grounded advice distilled from successful evolutionary mutations rather than general episodic or document memory.

Tool-augmented systems are relevant insofar as they show how code generation, execution, retrieval, and reflection can be organized into reusable workflows \citep{ren2025scientificintelligence,hasib2025processcentric,li2025deepcode}. Among them, OR-Agent is the closest comparison because it combines structured research loops with memory-based reflection, while SciDataCopilot similarly emphasizes reusable workflow generation \citep{liu2026oragent,rao2026scidatacopilot}. Other systems such as AI Scientist, Paper2Code, CodeScientist, and MLR-Copilot demonstrate broader executable research pipelines, but they are less specifically focused on persistent memory for evolutionary mutation and cross-run reuse \citep{lu2024aiscientist,paper2code2025,codescientist2025,li2024mlrcopilot}.

\section{Memory architecture}
\label{sec:memory-architecture}

GigaEvo follows a standard evolutionary loop: it selects parent
programs, asks an LLM-based mutation operator to modify them,
executes the resulting candidates, and stores their metrics. In the
baseline setting, each mutation sees only local run context such as
the parent code, current metrics, recent insights, and lineage.
This keeps the search procedure simple and controlled, but it also
means that useful tactics discovered in one run are not available in
later runs. EvoMem addresses this limitation by converting
successful mutation events into reusable, task-aware advice.

The design is intentionally split into two phases. The \emph{write
phase} runs after an evolution job finishes, extracts generalizable
ideas from the produced programs, and stores them in a memory
store. The \emph{read phase} runs before each new mutation, retrieves a
small set of relevant memory cards, and injects them into the mutation
prompt. This separation is important for efficiency: expensive
operations such as clustering, deduplication, and optional LLM-based
enrichment happen offline after the run, while mutation-time retrieval
remains a small bounded step.

\subsection{General pipeline}
\label{sec:general}

At a high level, a completed run produces a set of programs, a
post-run analysis stage extracts candidate ideas from those programs,
the memory layer normalizes and stores them, and future runs query this
store before proposing a new mutation. We use the following compact
notation to make the data flow explicit:
\begin{equation}
\label{eq:pipeline}
P_r \xrightarrow{\mathrm{extract}} C_r
    \xrightarrow{\mathrm{write}} M
    \xrightarrow{\mathrm{select}(p,\tau,\mu)} I_p
    \hookrightarrow \mathrm{mutation\ prompt}.
\end{equation}
Here $P_r$ is the set of programs from run $r$, $C_r$ is the set of
cards extracted from that run, $M$ is the persistent memory bank,
and $I_p$ is the small list of instructions selected for parent
program $p$ on task $\tau$ with metric description $\mu$.

Conceptually, the architecture has three parts. First, a post-run
analysis stage reviews completed runs and produces structured records
of candidate ideas together with provenance information. Second, a
persistent memory layer stores these records, merges near-duplicates,
and maintains retrieval indexes. Third, a retrieval layer queries this
store at
mutation time. In control runs, this layer is simply inactive, so the
same evolutionary pipeline can be used both with and without memory.

\subsection{Insights extraction and processing}
\label{sec:insights}

The write phase begins from mutation records attached to evolved
programs. During the run, the mutation process records a list of
changes; each change has a short description and, when available, a
rationale for why it was introduced. After the run, the analysis stage
collects the completed programs, filters out root programs and invalid
outputs, and converts the remaining candidates into a normalized set of
records containing fitness, generation, parentage, task context,
mutation strategy, code, and improvement descriptions.

There are two analysis modes. The default mode compares new
improvements against a working collection of previously extracted
ideas. For each program, an LLM decides whether the recorded
improvements correspond to genuinely new ideas, revisions of existing
ideas, or simple reformulations of earlier descriptions. The working
collection is then updated while preserving provenance: source program
references are appended, the latest generation is updated, older
descriptions are kept as aliases when wording changes, and new
rationales are appended to the explanation history.

The fast analyzer is used when a run contains many similar improvement
descriptions. It first groups semantically related candidates using
DBSCAN over embedding space, then asks an LLM to refine ambiguous
multi-member clusters. This produces a smaller set of reusable idea
candidates while preserving the source programs and rationales as
provenance. Cross-run consolidation is deferred to the memory insertion
stage, where new ideas are compared against the existing corpus.

After analysis, the active memory is enriched with task summaries,
keywords, and short explanation summaries. If memory-usage tracking is
enabled, the system also uses the selected memory-card identities
recorded on parent programs to estimate downstream impact. For a child
program, the recorded contribution of a selected memory item is
\[
\Delta f = f(\mathrm{child}) - \max_{p \in \mathrm{parents}} f(p),
\]
computed only when both child and parent fitness values are valid, i.e., when both passed a task--specific validity check.
The stored usage statistics keep per-task counts, all observed deltas,
and the median delta. Median aggregation is used because evolutionary
fitness changes are typically sparse and heavy-tailed.

\subsection{Memory generation}
\label{sec:generation}

The write pipeline stores two kinds of entries. The first represents
an abstract idea or tactic, including its description, task context,
explanations, provenance, and usage statistics. The second represents
a high-performing program together with its fitness, code, task
context, and links to the ideas associated with it. Before insertion,
all incoming data are normalized into these structured forms. This
normalization step is necessary because the write pipeline consumes
heterogeneous inputs---LLM outputs, post-run analysis logs, and
historical dumps---so missing fields and legacy aliases must be
handled consistently.

Not every extracted idea is promoted into long-term memory with equal
status. In addition to the full collection of extracted ideas, the
system maintains a smaller snapshot of high-value ideas intended to
capture the most reusable findings from a run. After the post-run
analysis stage finishes, an origin-analysis pass scores each idea by
both how it first appeared in the evolutionary lineage and how it
shaped later descendants. For each introduction event, the analysis
compares the child program with its strongest parent, with sibling
programs generated from the same parent context, and with downstream
descendants that inherit the idea.

The promotion rule is deliberately conservative rather than tuned to a
particular benchmark. An idea is favored when it has an identifiable
introduction point, improves over its strongest available parent,
rarely produces worse descendants, and is supported by at least one
additional signal such as recurrence across nearby programs, favorable
comparison with siblings generated from the same context, or spread
into later elite lineages. This filter biases memory toward ideas with
lineage-level evidence of usefulness, rather than toward every change
described by the LLM.

Once this curated set has been formed, the write pipeline combines the
full idea collection, the filtered high-value snapshot, optional
high-performing program exemplars, and usage updates. New records are
then inserted into persistent memory under a conservative policy:

\begin{enumerate}
    \item repeated observations of the same memory entity update the
    existing record and extend its provenance;
    \item concrete program exemplars are stored without semantic
    merging, because they are tied to specific artifacts and fitness
    values;
    \item abstract idea cards are merged only when they describe the
    same underlying mechanism, not merely the same source task.
\end{enumerate}

To identify possible duplicates, EvoMem compares each incoming idea
against multiple textual views of existing cards, including their
mechanism descriptions, task summaries, and explanations. Candidate
scores are combined by a weighted sum,
\[
\mathrm{score}(c)=\sum_{q \in Q} w_q s_q(c),
\]
where $s_q(c)$ is the retriever score for candidate card $c$ under
query view $q$. The default weights are largest on the description
and description--explanation views, because these are the best
signals for whether two cards describe the same mechanism rather
than merely the same benchmark.

The highest-scoring candidates are passed to an LLM decision policy
that chooses whether to store the new idea, discard it as redundant, or
merge it with an existing entry. Ambiguous cases are retained as new
cards, favoring recall over accidental loss of useful knowledge.
Deduplication is therefore an optimization of corpus quality, not a
precondition for preserving a potentially useful idea.

The resulting memory corpus is stored persistently together with
provenance and usage statistics, and retrieval indexes are refreshed so
that later runs can query the updated memory bank.

\subsection{Memory retrieval and usage}
\label{sec:retrieval}

Before each mutation, EvoMem optionally augments the candidate context
with retrieved memory. In control runs this augmentation is empty; in
memory-enabled runs, a retrieval module selects a bounded set of
relevant memory cards. This keeps the evolutionary operators fixed
across conditions: parent selection, mutation generation, validation,
and scoring are unchanged, while memory enters only as an additional
advice channel.

The retrieval query is derived from the current task description,
metric descriptions, mutation mode, parent program, and local mutation
context. EvoMem then selects a small number of concise, actionable
ideas. Thus, the intervention is localized: the current program state
is used to retrieve advice, but program generation and evaluation
remain unchanged.

Before retrieving individual ideas, EvoMem performs an additional
task-scope filtering step. An LLM is shown the current task together
with the task descriptions present in memory and selects the source
tasks that are potentially relevant. Subsequent retrieval is then
performed over ideas associated with those selected tasks, rather than
over the entire memory corpus. This keeps retrieval focused while
still allowing transfer from related but non-identical problems.

Within the selected task scope, EvoMem ranks memory cards using lexical
and embedding-based similarity over several semantic views: the card's
mechanism description, the source task context, the explanation of why
the idea helped, and selected compositions of these fields. This
decomposition is important because a useful memory card may match the
present mutation through different signals: the mechanism it describes,
the task constraints under which it was observed, or the explanation of
why it improved performance.

The selection policy merges evidence from multiple retrieval channels
and may issue a focused follow-up query when the initial evidence is
insufficient. Both the number of retrieval iterations and the number of
returned cards are fixed in advance, so retrieval remains a bounded
mutation-time operation rather than an open-ended research process.

The retrieval step returns two coupled outputs: a concise advice block
inserted into the mutation prompt, and the identities of the selected
memory cards. These identities are recorded with the resulting
candidate so that, after evaluation, the system can estimate
associations between memory use and downstream fitness changes. This
record is what enables the usage-tracking pass in the next write phase.

Finally, the retrieved text is inserted into the mutation prompt as a
dedicated memory section alongside metrics, recent insights, lineage,
evolutionary statistics, and any task-specific artifact context. The
mutation agent therefore sees memory as advice rather than as a hard
constraint. It may reuse the retrieved tactics when they are relevant,
while every candidate program is still evaluated by the same
validators and metrics as in non-memory runs.

\subsection{Design properties}

Three design choices are especially important for experiments. First,
memory is \emph{optional}: control and treatment runs differ only in
whether the memory advice channel is populated. Second, retrieval is
\emph{bounded}: mutation-time memory access is limited by fixed item
and iteration budgets. Third, the system is \emph{auditable}: selected
memory cards are recorded with the candidates that used them, and
post-run analysis can trace those links back to the relevant source
programs. Together, these properties make it easier to attribute
measured gains to the memory mechanism rather than to unrelated changes
elsewhere in the pipeline.

This design is intended to limit negative transfer. EvoMem does not replace the
core evolutionary process with memory replay: parent selection,
mutation generation, validation, evaluation, and population updates
remain the same as in the baseline system. Memory enters only as a
bounded suggestion channel inside the mutation context, and retrieved
ideas are not enforced as constraints. As a result, the search can
ignore irrelevant memories, continue proposing novel mutations, and
select against candidates that overfit to stale or task-mismatched
advice. The memory layer therefore aims to bias exploration toward previously
useful directions without eliminating the discovery of new strategies.

\section{Methodology}
For memory-mechanism evaluation, we build memories from prior runs but exclude same-benchmark memories at test time. The memory-generation set spans geometry, question answering, and code optimization. It includes Circle Packing (26), Heilbronn-style point placement, Kissing Number (11D), HotpotQA \citep{yang-etal-2018-hotpotqa}, HoVer \citep{jiang-etal-2020-hover}, GSM8K \citep{cobbe2021trainingverifiers}, selected AlgoTune tasks \{QP, Markowitz, LP-Box, Chebyshev Center, Feedback Controller, LQR, Lyapunov\} \citep{press2025algotune}, and selected KernelBench kernels \{L1 P56, L1 P82, L1 P86, L1 P87, L2 P57\} \citep{ouyang2025kernelbenchllmswriteefficient}. Only part of this set is drawn from established benchmark suites, including AlgoTune for scientific-code tuning and KernelBench for GPU-kernel optimization; the remaining tasks cover geometric optimization and reasoning settings used to probe transfer more broadly.

The evaluation set contains tasks from the same broad families but is treated as a separate testing set for cross-task memory reuse. It includes Circle Packing (26 and 32), Hexagon Packing, Heilbronn-style point placement, Kissing Number (12D), HoVer, HotpotQA, AlgoTune Power Control, AlgoTune Kalman, and KernelBench kernels including L1 P66 and L3 P21. Circle packing and hexagon packing ask the system to arrange shapes as densely as possible inside a bounded region. Heilbronn-style tasks optimize point placements by maximizing the minimum triangle area induced by triples of points. Kissing Number evaluates dense high-dimensional sphere configurations. HotpotQA and HoVer test multi-hop question answering and evidence integration, with QA performance measured by exact match on a fixed test set of 200 samples for each benchmark. AlgoTune evaluation tasks measure speedups of scientific and algorithmic routines under correctness constraints, and KernelBench evaluation tasks measure low-level GPU-kernel optimization while preserving correctness.

All components of the GigaEvo mutation framework used Gemini 3 Flash as the LLM backbone \citep{gemini3flash2025}. Prompt-evolution experiments with LLM-based evaluation or generation components, including HotpotQA, HoVer, and GSM8K, used Qwen3-8B as the tested LLM \citep{yang2025qwen3}.

For reproducibility, we provide the full EvoMem implementation code with the submission. Unless otherwise noted, LLM calls used temperature $0.7$, the MAP-Elites population used a single island, memory retrieval returned at most $k=3$ cards per mutation, and embedding-based retrieval used the all-MiniLM-L6-v2 embedding model. Across the reported experiments, the evolution phase used on average 7{,}192{,}260 input tokens and 892{,}674 output tokens, while the post-run memory-generation phase used on average 683{,}470 input tokens and 25{,}225 output tokens. At the idea level, each memory bank session extracted $99.71 \pm 38.68$ generated ideas from one completed evolution run, and selected $12.76 \pm 7.51$ ideas for insertion into the memory bank after filtering and processing.

We build the memory bank from the memory-generation set and then test EvoMem on the evaluation set. When evaluating a benchmark, we exclude all memories produced from that benchmark from the memory bank. Thus, the memory-enabled run can only retrieve memories from other benchmarks, including related tasks, and never from the benchmark currently being tested. For each evaluation benchmark, we compare a baseline condition without memory against the same evolutionary pipeline with retrieved memories enabled. The comparison measures whether reusable ideas extracted from prior tasks improve final performance or search speed on held-out tasks. 

\section{Results}

\begin{figure}[t]
    \centering
    \begin{subfigure}[t]{0.48\textwidth}
        \centering
        \includegraphics[width=\linewidth]{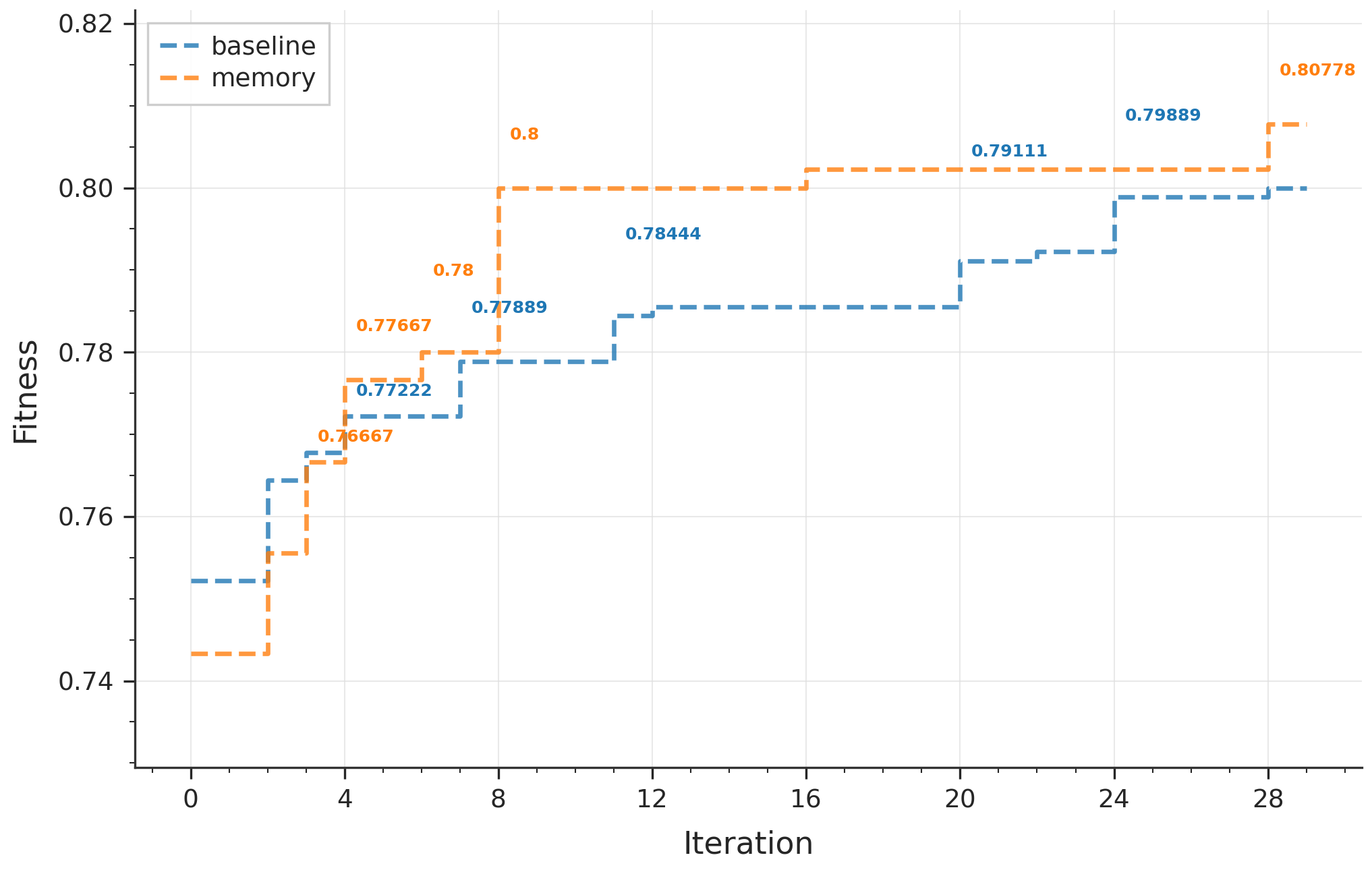}
        \caption{HoVer.}
    \end{subfigure}\hfill
    \begin{subfigure}[t]{0.48\textwidth}
        \centering
        \includegraphics[width=\linewidth]{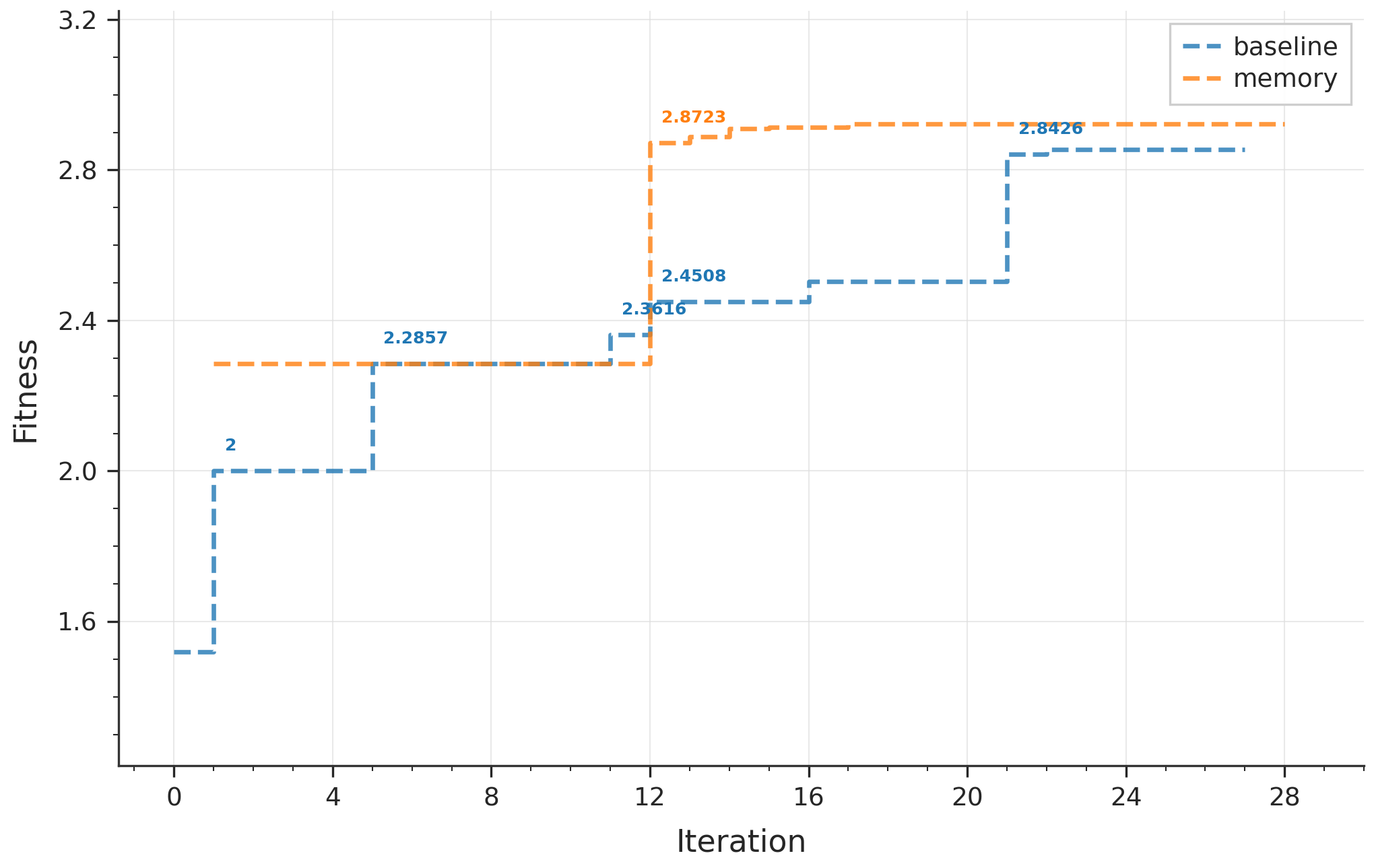}
        \caption{Circle Packing n=32.}
    \end{subfigure}

    \vspace{0.5em}

    \begin{subfigure}[t]{0.48\textwidth}
        \centering
        \includegraphics[width=\linewidth]{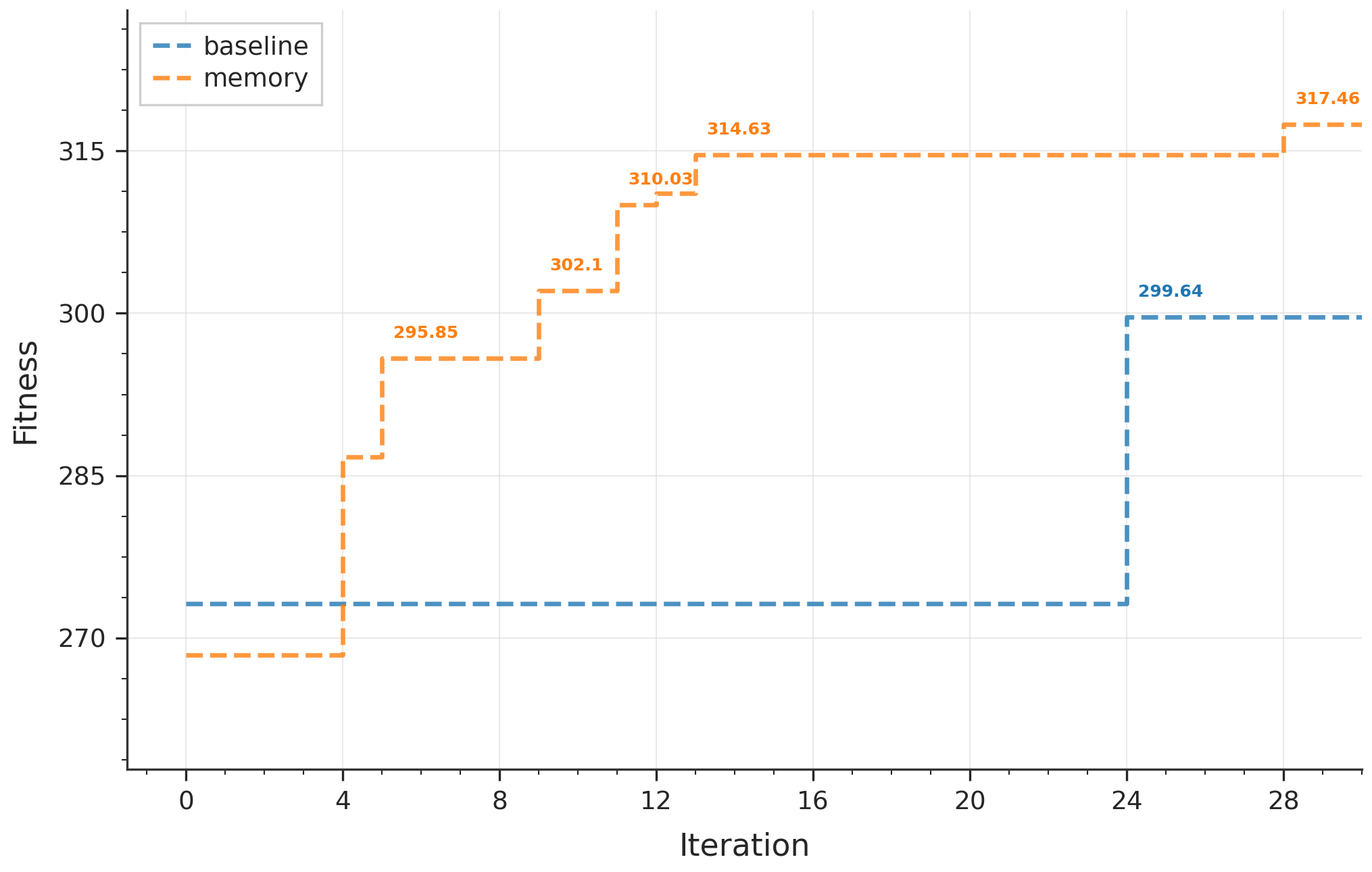}
        \caption{Kalman filter.}
    \end{subfigure}\hfill
    \begin{subfigure}[t]{0.48\textwidth}
        \centering
        \includegraphics[width=\linewidth]{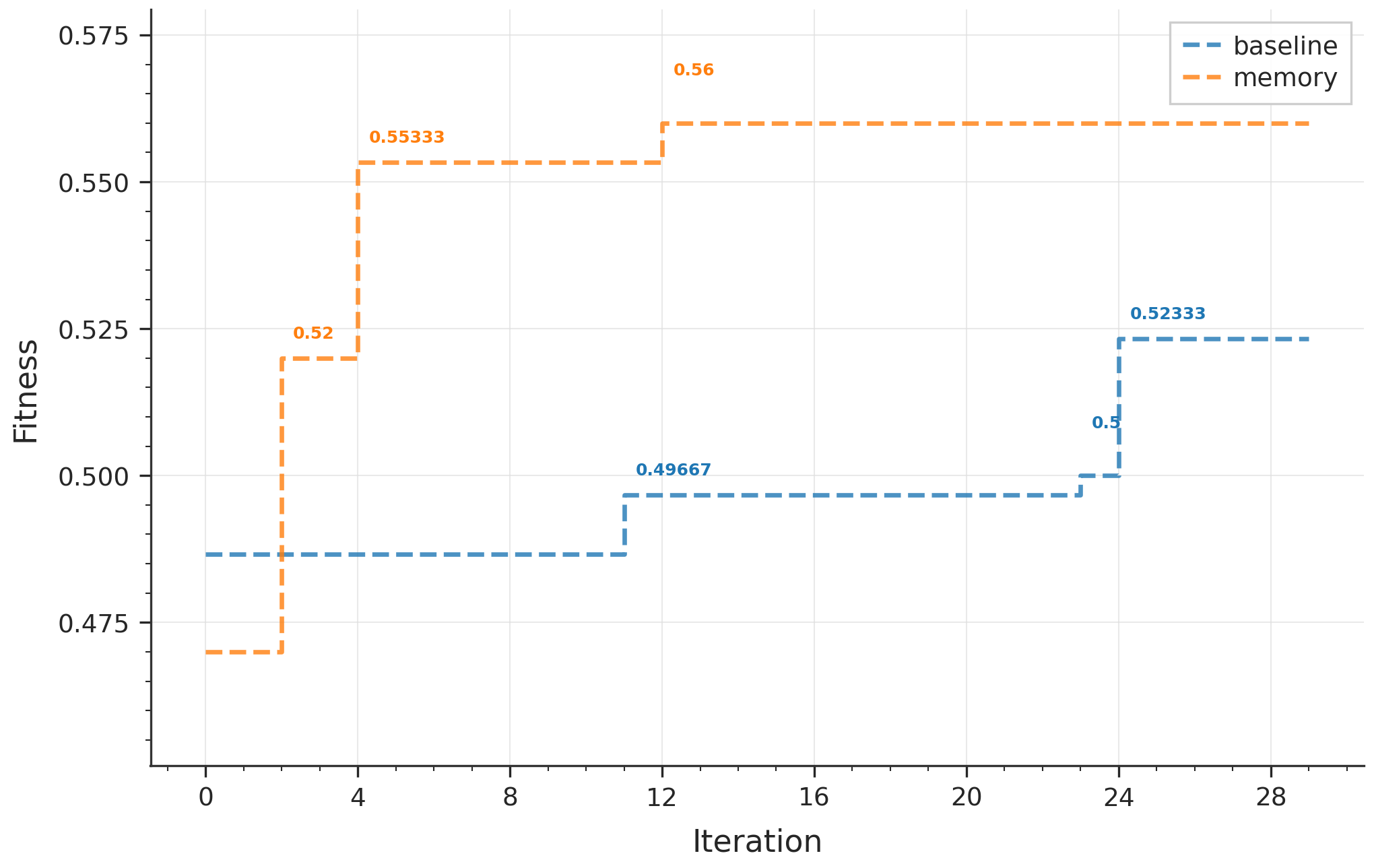}
        \caption{HotpotQA.}
    \end{subfigure}

    \caption{Representative transfer cases across reasoning and optimization benchmarks. Each panel compares baseline and memory-enabled behavior for a different task.}
    \label{fig:four_images}
\end{figure}

We evaluate EvoMem on cross-problem transfer experiments spanning geometric optimization, multi-hop question answering, GPU kernel optimization, and scientific-code tuning. In each setting, we compare a baseline target-benchmark run against a memory-enabled run whose memory bank excludes memories from that target benchmark. The results are reported in Table~\ref{tab:results-summary}. Figure \ref{fig:four_images} shows representative cases in which memory is used on different prompt and algorithmic optimization tasks.

Search speedup is computed as the ratio between the number of evaluated
candidates needed by the baseline run and the memory-enabled run to reach the
baseline run's final best score.

\begin{table}[!htp]\centering\small
\caption{Summary of target-metric relative gains and search-speed ratios for memory-enabled runs across evaluated benchmarks. A speedup greater than 1 indicates that the memory-enabled run reached the baseline run's final best score using fewer evaluated candidates; values below 1 indicate slower convergence. Relative gain is calculated separately for each matched pair of runs as \(g_i=(F_i^{\mathrm{memory}}-F_i^{\mathrm{baseline}})/|F_i^{\mathrm{baseline}}|\). The reported average, minimum, and maximum gains summarize the resulting pairwise \(g_i\) values.
\vspace{0.1em}
}
\label{tab:results-summary}
\resizebox{\linewidth}{!}{%
\begin{tabular}{lrrrrrrr}\toprule
\textbf{Benchmark} &\textbf{Average gain} &\textbf{Min gain} &\textbf{Max gain} &\textbf{Average speedup} &\textbf{Min speedup} &\textbf{Max speedup} \\\midrule
Circle Packing (32) &5.21\% &0.48\% &9.29\% &9.26 &3.29 &22.00 \\
Circle Packing (26) &5.88\% &0.28\% &17.13\% &5.06 &0.31 &9.00 \\
Heilbronn &3.86\% &0.00\% &8.55\% &6.01 &0.31 &16.17 \\
Kissing Number (12D) &0.00\% &0.00\% &0.00\% &6.78 &6.17 &7.40 \\
Hexagon Packing &3.69\% &0.38\% &6.93\% &5.75 &0.75 &12.50 \\
HoVer &2.42\% &2.28\% &2.56\% &2.71 &1.92 &3.50 \\
AlgoTune &7.90\% &6.22\% &9.57\% &8.58 &0.82 &16.33 \\
KernelBench &16.89\% &0.00\% &65.93\% &3.44 &0.05 &9.00 \\
HotpotQA &11.79\% &5.16\% &27.41\% &5.75 &1.27 &12.00 \\
\midrule
\textbf{Average across benchmarks} &\textbf{6.40\%} &\textbf{1.64\%} &\textbf{16.37\%} &\textbf{5.93} &\textbf{1.65} &\textbf{11.99} \\
\bottomrule
\end{tabular}}
\end{table}

Table~\ref{tab:results-summary} shows that EvoMem is associated with better target metrics, faster search, or both across several benchmark families. The strongest speedups appear in Heilbronn, Kissing Number, HotpotQA, and Circle Packing, while AlgoTune and KernelBench show positive mean gains with higher variability. Kissing Number illustrates an acceleration-only case: final quality is unchanged on average, but memory-enabled runs reach the baseline best score earlier. Absolute target-metric values are reported in Appendix~\ref{app:benchmark-absolute-values}.

\subsection{Examples of retrieved memory reuse}
\label{sec:memory-reuse-examples}

To make the effect of memory more concrete,
Table~\ref{tab:memory-reuse-examples} summarizes representative cases
where retrieved memories were incorporated into successful mutations.
These examples illustrate that the stored memories were not limited to
copying task-specific code. Instead, they often expressed reusable
execution strategies, search heuristics, or reasoning patterns that
could be adapted to new target tasks.

Across these cases, retrieved memories usually encoded transferable execution
ideas---such as caching repeated structure, exploiting sparsity, reducing
convergence checks, or limiting custom kernels to profitable fusion---rather
than narrow task-specific tricks.

\begin{table}[!htbp]
\centering
\footnotesize
\caption{Representative examples of memory ideas reused during evolution.}
\label{tab:memory-reuse-examples}
\resizebox{\textwidth}{!}{%
\begin{tabular}{p{0.17\linewidth}p{0.18\linewidth}p{0.29\linewidth}p{0.27\linewidth}}
\toprule
\textbf{Target task} & \textbf{Source or inferred source} & \textbf{Retrieved memory idea} & \textbf{How it was used} \\
\midrule
Circle Packing & Heilbronn-style point placement &
Use annealed force-directed repulsion and low-discrepancy initialization to improve geometric dispersion. &
Replaced the static grid with force-directed motion and adaptive radius expansion. \\

Hexagon Packing & Circle/geometric optimization &
Use basin hopping, simulated annealing, and temperature-scaled repulsive forces to escape local packing optima. &
Added hex-lattice starts, Gaussian jitter, and boundary-aware relaxation. \\

HotpotQA & Multi-hop QA traces &
Extract only claim-relevant facts, preserve intermediate entities, and maintain query-refinement continuity across steps. &
Added entity extraction and shorter final synthesis to reduce context drift. \\

HoVer & GSM8K / arithmetic reasoning &
Make latent relationships explicit and verify them with concrete checks rather than generic reasoning instructions. &
Added explicit reasoning questions, missing-fact lists, and targeted query examples. \\

KernelBench L1 P66 & Kernel optimization traces &
Avoid over-customized kernels when vendor primitives are faster; use custom Triton kernels only for profitable fusion. &
Used \texttt{F.conv3d} for Conv3D and kept Triton only for fused bias/ReLU. \\

Power Control & Markowitz / QP-style optimization &
Precompute invariant matrix scaling, use in-place linear algebra, and reduce convergence-check overhead. &
Precomputed scaled coefficients, reused buffers, and checked convergence less often. \\

Battery Scheduling & LP-Box / Chebyshev Center / QP-style optimization &
Cache repeated constraint structure, use sparse matrices for structured LPs, tune HiGHS options, and avoid repeated conversions. &
Cached horizon templates, used CSR matrices, tuned HiGHS, and delayed conversions. \\
\bottomrule
\end{tabular}}
\end{table}
\newpage
\section{Memory Quality and Utilization}
\label{sec:memory-quality-utilization}

The benchmark results show whether memory-enabled evolution improves an
end metric, but they do not by themselves show how broadly the bank is queried
or whether the retrieved advice is actually incorporated into generated code.
We therefore complement the outcome analysis with two mechanism-level audits:
memory bank coverage in the transfer-target experiments and an acceptance audit
that compares the default retrieval mechanism with random memory selection.

\subsection{Memory utilization across transfer targets}

The memory-usage reports associate every generated program with the identifiers
of the memory cards selected for its mutation. We count a bank idea as
\emph{retrieved} when its identifier occurs at least once in a candidate's metadata and matches a card in the bank
snapshot loaded for that experiment. This is a breadth measure and does not
increase when the same card is retrieved repeatedly.

\begin{figure}[t]
    \centering
    \includegraphics[width=\linewidth]{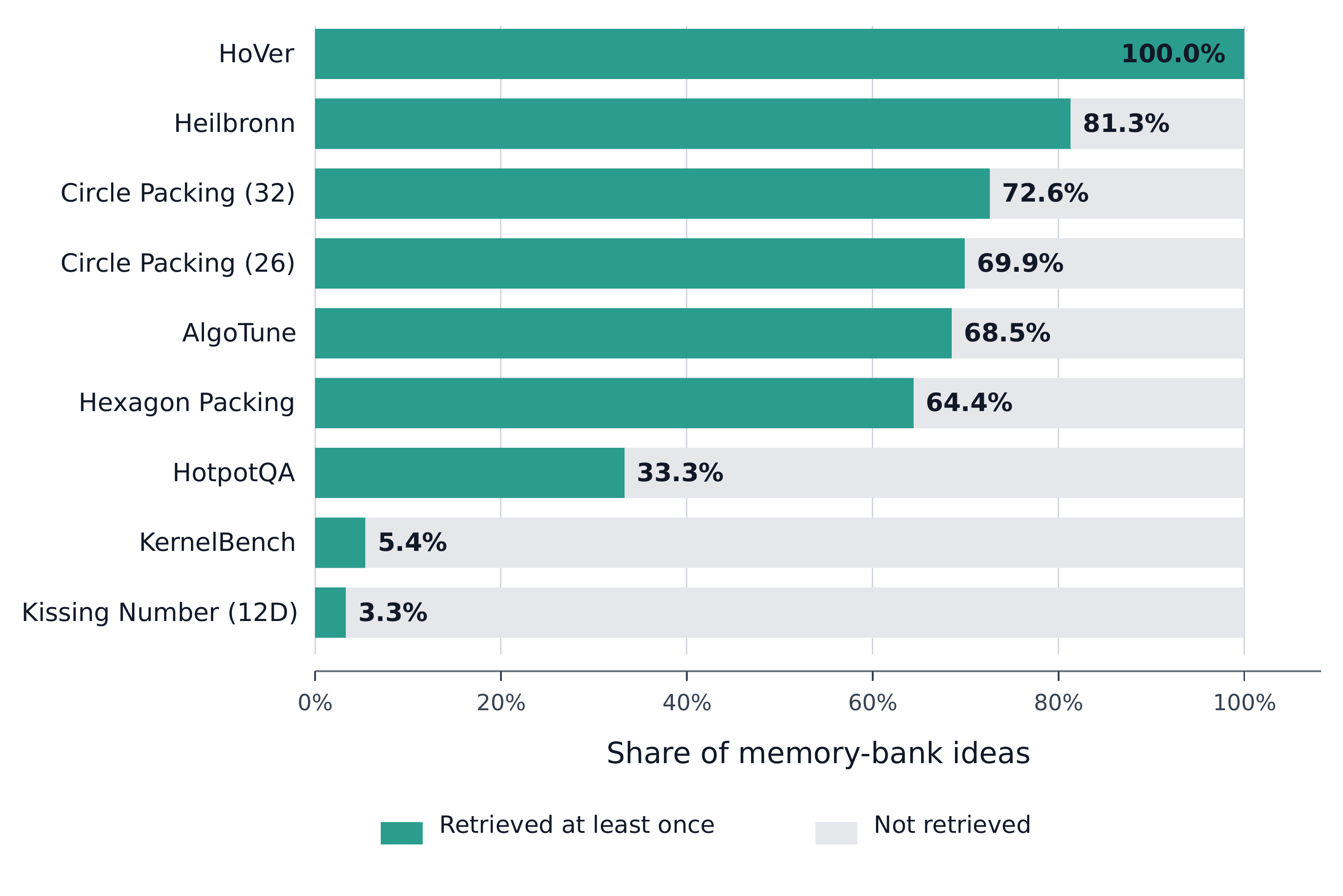}
    \caption{Breadth of memory bank use in the completed transfer experiments for which usage reports were available. Each bar gives the percentage of the available memory bank retrieved at least once, with the two KernelBench targets grouped into one benchmark-family bar. This measures retrieval coverage, not retrieval frequency or implementation.}
    \label{fig:memory-bank-utilization}
\end{figure}

Figure~\ref{fig:memory-bank-utilization} shows substantial variation in how much
of each bank was explored. The only benchmark with full memory bank coverage was HoVer.
The completed Heilbronn, Circle Packing (26), and
Hexagon Packing reports retrieved 81.3\%, 69.9\%, and 64.4\% of their
available bank ideas, respectively. HotpotQA retrieved 33.3\%.
The KernelBench coverage was 5.4\%, while Kissing Number (12D) showed lowest coverage, at 3.3\%.

\subsection{Retrieval quality and idea acceptance}

Coverage measures whether an idea was selected, but selection alone does not
establish that the mutation agent followed the advice. We therefore compare
EvoMem's default relevance-based retrieval with random sampling from the same
memory bank. Each selected program--idea pair was examined by an automated
parent--child code audit. An idea was counted as \emph{accepted} only when the
child newly introduced or materially strengthened the selected idea relative to
its parents. The acceptance rate is computed over decided checks,
\(\text{accepted}/(\text{accepted}+\text{not accepted})\); evaluator errors are
excluded from this denominator.

Across the two retrieval-quality audit groups, default retrieval achieved a 4.05\% acceptance rate. On the other hand, random memory
achieved only 0.25\%. Thus,
on average, the retrieval mechanism
outperforms random memory selection: a retrieved idea was 16.2 times as likely
to be incorporated into the child program, an absolute increase of 3.80
percentage points. The comparison is not driven by one audit group: the rates
were 6.23\% versus 0.30\% in the first group and 1.88\% versus 0.20\% in the
second, and default retrieval had the higher rate in every matched comparison.
Both conditions drew from the same memory bank.

\begin{figure}[t]
    \centering
    \includegraphics[width=0.92\linewidth]{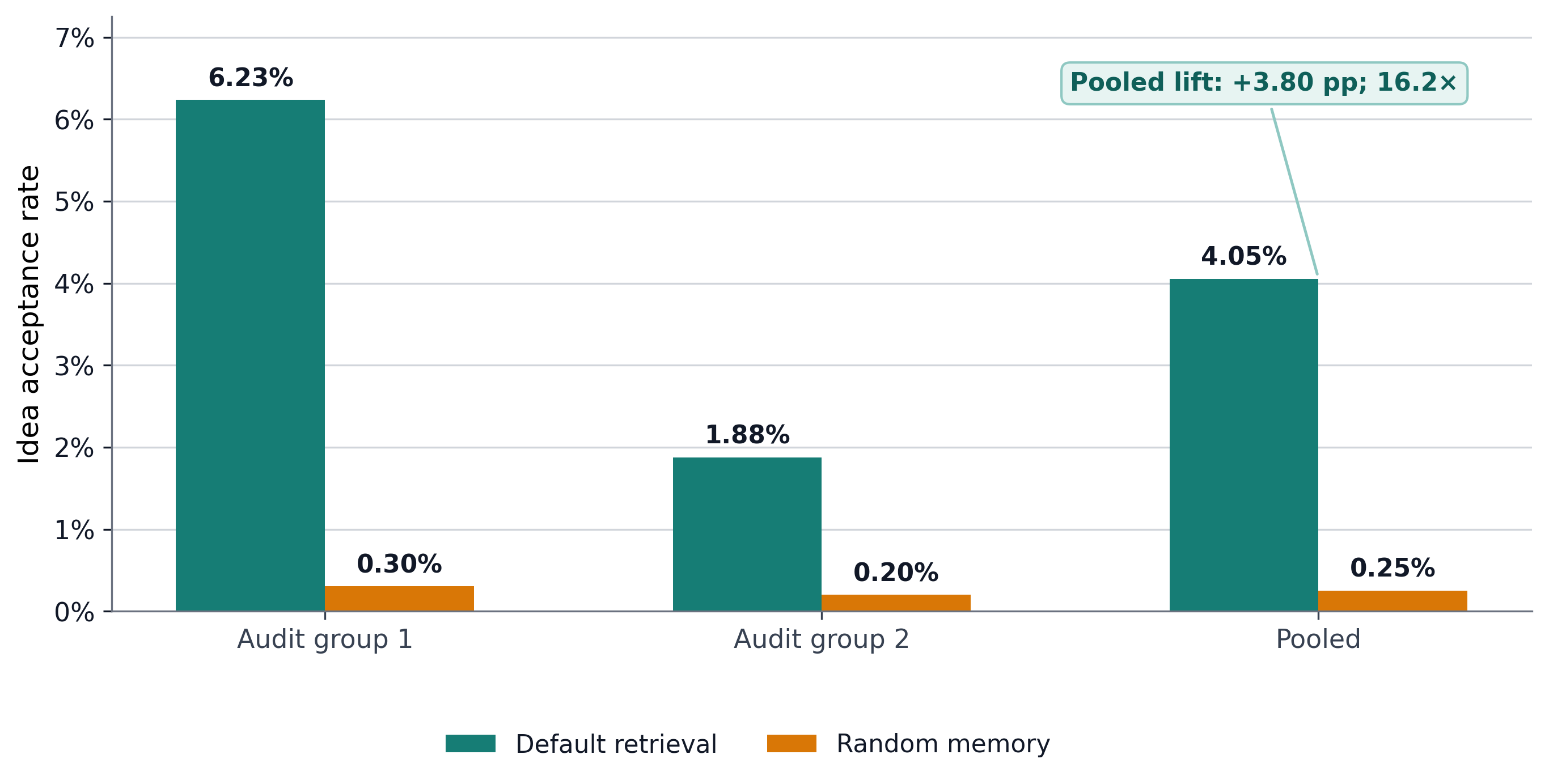}
    \caption{Idea-acceptance rates under default relevance-based retrieval and random memory selection. The pooled bars combine both audit groups. Rates use decided program--idea checks.}
    \label{fig:memory-quality-acceptance}
\end{figure}

The program-level view provides the same qualitative result. At least one
retrieved idea was accepted in 136 of 1{,}442 audited programs (9.43\%), compared
with 10 of 1{,}451 programs (0.69\%) under random memory, a 13.7-fold difference.
Together with Figure~\ref{fig:memory-quality-acceptance}, this shows that the
retrieval layer does more than expose the mutation model to arbitrary prior
ideas: its relevance filtering selects advice that is substantially more likely
to be realized in code. This mechanism-level evidence complements, rather than
replaces, the end-to-end fitness and search-speed comparisons above.

\section{Conclusion}

We introduced EvoMem, a persistent memory architecture for LLM-based
evolutionary code optimization. EvoMem converts successful mutation events
from previous runs into structured, auditable memory cards and retrieves a
bounded set of relevant cards during later mutations. This lets an
evolutionary search process reuse previously discovered tactics without
replaying old programs directly or changing the evaluator, parent-selection
policy, or fitness computation.

Across the evaluated geometry, reasoning, scientific-code, and GPU-kernel
benchmarks, EvoMem is associated with improved efficiency or final target
metrics in most of the tested settings. The results suggest that some useful
mutation knowledge is more general than the single program in which it first
appeared: strategies discovered in one run can help guide exploration in
related and sometimes substantially different domains. These gains provide
initial support for the central hypothesis of this work, namely that
LLM-driven evolution can benefit from persistent cross-run memory because it
may reduce redundant rediscovery and bias search toward previously productive
directions while preserving evaluator-based selection.

More broadly, EvoMem suggests that memory should be treated as a first-class
component of evolutionary coding systems. As evolutionary agents become more
capable, the artifacts produced by a run include not only final code but also
reusable design decisions, optimization heuristics, and failure-avoidance
patterns. Capturing these artifacts in a compact and retrievable form may
provide a practical path toward evolutionary systems that accumulate
experience across tasks instead of starting each optimization process from
scratch.

\section{Limitations and Future Work}

The benefits of EvoMem vary across runs, as reflected in the wide range between the minimum and maximum gains and speedups observed for several benchmarks. This variance is partly inherent to
evolutionary search: useful exploration often requires high-temperature LLM
sampling, so repeated runs can follow different trajectories. At the same time,
too little diversity, or overly rigid use of retrieved memory, can cause
premature convergence. EvoMem is also sensitive to the quality of its memory
pipeline, including idea extraction, clustering, summarization, retrieval, and
their associated hyperparameters. Poorly formed or poorly retrieved memories can
reduce the relevance of mutation advice and limit downstream gains.

The present study also uses a relatively small memory bank, a fixed memory
influence during mutation, and a simple code-evolution setting. Future work
should scale memory across more runs and domains, adapt memory reliance to the
task and population state, and study whether persistent memory can reduce the
cost of repeated LLM calls, execution, validation, and selection. Another
important direction is repository-scale evolution, where memories may capture
project conventions, interface constraints, recurring bug patterns, and
cross-file optimization strategies.

Because EvoMem is evaluated inside full evolutionary optimization loops,
component-level ablations are substantially more expensive than ordinary prompt
or retrieval ablations: each setting changes the search trajectory and requires
a fresh set of evaluator-in-the-loop runs. We therefore treat the present
experiments as an end-to-end evaluation of the memory mechanism rather than a
complete decomposition of all design choices.

\bibliography{main}

\newpage
\appendix

\section{Ablation Studies}
\label{sec:ablation-studies}

We conduct a controlled ablation study on Circle Packing (32) to separate the
effect of the complete EvoMem mechanism from simpler forms of prompt and memory
augmentation. Circle packing provides a well-understood optimization objective
with a known reference solution. Because each condition requires a fresh
evaluator-in-the-loop evolutionary run, we allocate the experimental budget to
five independent trials per condition on this task rather than to fewer trials
spread across several benchmarks. We compare the full default EvoMem
configuration with no memory, random memory, generic advice, memory without task
filtering, and retrieval-only memory without the complete re-ranking procedure.

Near-optimal improvements are difficult to interpret from raw fitness alone:
scores around $2.8$ are comparatively easy to reach, whereas small absolute
gains become increasingly consequential near the best-known fitness
$f^*=2.9370$. We therefore also report the percentage of the remaining target
gap closed. For a fitness value $f$, this quantity is
\begin{equation}
    \operatorname{GapClosed}(f)
    = 100\,\frac{f-\bar{f}_{\mathrm{base}}}
    {f^*-\bar{f}_{\mathrm{base}}},
    \qquad \bar{f}_{\mathrm{base}}=2.8151.
\end{equation}
Under this normalization, the no-memory mean is $0\%$ and the best-known
fitness is $100\%$. Standard deviations are transformed by the same fixed
scale; consequently, the baseline can have zero mean gap closure while retaining
a nonzero standard deviation.

\begin{table}[!htbp]
\centering
\small
\caption{Circle Packing (32) ablation results over five independent trials per
condition. Fitness and target-gap closure are reported as mean $\pm$ standard
deviation; best and worst denote the extrema across the five trials. Ablation runs use an independent set of trials and are not included in the data reported in Table~\ref{tab:results-summary}.}
\label{tab:ablation-circle-packing-32}
\resizebox{\linewidth}{!}{%
\begin{tabular}{lrrrr}
\toprule
\textbf{Condition} & \textbf{Fitness (mean $\pm$ SD)} & \textbf{Best run} &
\textbf{Worst run} & \textbf{Target gap closed (mean $\pm$ SD)} \\
\midrule
Baseline (no memory) & $2.8151 \pm 0.0683$ & $2.9061$ & $2.7388$ & $0.0\% \pm 56.0\%$ \\
Random memory & $2.8219 \pm 0.0681$ & $2.8900$ & $2.7538$ & $5.6\% \pm 55.9\%$ \\
Generic advice & $2.8805 \pm 0.0530$ & $2.9228$ & $2.7877$ & $53.7\% \pm 43.5\%$ \\
No task filtering & $2.8714 \pm 0.1148$ & $2.9352$ & $2.6683$ & $46.2\% \pm 94.2\%$ \\
Retrieval-only memory & $2.8379 \pm 0.2210$ & $2.9370$ & $2.4426$ & $18.7\% \pm 181.3\%$ \\
Full EvoMem & $2.9338 \pm 0.0044$ & $2.9370$ & $2.9278$ & $97.4\% \pm 3.6\%$ \\
\bottomrule
\end{tabular}}
\end{table}

Table~\ref{tab:ablation-circle-packing-32} shows that arbitrary memory content
or retrieval alone does not reliably reproduce EvoMem's gains. Random memory
closes only $5.6\%$ of the target gap on average, and retrieval-only memory
closes $18.7\%$ while exhibiting the largest variance; the one run that reaches
the best-known fitness is therefore not representative of its overall
performance. Generic advice
and removing task filtering perform better on average, closing $53.7\%$ and
$46.2\%$ of the gap, respectively, but both remain below the full method, and
the no-task-filtering condition is particularly unstable.

In contrast, full EvoMem attains a mean fitness of $2.9338 \pm 0.0044$ and
closes $97.4\% \pm 3.6\%$ of the target gap. Even its worst run, at $2.9278$,
exceeds the no-memory baseline mean of $2.8151$. The simultaneous improvement
in average fitness and run-to-run consistency indicates that the strongest and
most reliable gains arise from the complete mechanism---constructing reusable
knowledge, selecting task-relevant memories, and integrating them into the
evolutionary process---rather than from additional context or retrieval
machinery alone.

\section{Lineage dynamics of memory use}
\label{sec:memory-lineage-dynamics}

The run logs also allow a finer-grained observational analysis inside
memory-enabled evolution. For each candidate, we mark whether the mutation
directly used retrieved memories and whether the candidate belongs to a
descendant lineage of any previous memory-use event. We then compare these
groups within the logs that contain both memory and non-memory post-root
mutations. Because the underlying tasks use different objective scales, the
analysis reports within-run objective ranks among valid post-root programs,
where larger ranks are better. These statistics are associative rather than
randomized causal estimates, but they test whether memory-guided variants are
preferentially retained by the evolutionary process.

\begin{table}[H]
\centering
\small
\caption{Lineage-level dynamics of retrieved memory usage. Objective values are converted to within-run ranks among valid post-root programs, where 1 is the best program in that run. Comparisons are made within the evolution logs between memory-associated programs and contemporaneous non-memory programs.}
\label{tab:memory-lineage-dynamics}
\resizebox{\linewidth}{!}{%
\begin{tabular}{lccc}
\toprule
\textbf{Statistic} & \textbf{Memory-associated} & \textbf{Comparator} & \textbf{Effect} \\
\midrule
Valid direct mutations & 47.9\% & 40.2\% & +7.7 pp \\
Direct children per mutation & 0.87 & 0.73 & 1.20$\times$ \\
Descendants per mutation & 2.57 & 2.21 & 1.16$\times$ \\
Mean objective rank of valid lineage members & 0.504 & 0.429 & matched +0.075 \\
Best lineage objective rank & 0.997 & 0.667 & +0.330 \\
\bottomrule
\end{tabular}}
\end{table}

Table~\ref{tab:memory-lineage-dynamics} reports that direct memory-use mutations
are more likely to produce valid candidates and have more reproductive output
than mutations without retrieved memory. The descendant view is also positive:
programs in memory-descended lineages have higher average objective rank, and
the strongest memory-descended lineages reach higher average best objective
ranks than the strongest non-memory lineages. This is consistent with the
intended role of EvoMem as a soft evolutionary bias: retrieved advice can be
used by mutations whose descendants remain competitive under selection.

\newpage

\section{Memory Card Schema and Algorithms}
\label{app:memory-schema}

\subsection{Memory card schema}
\label{app:memory-card-schema}

The schema below summarizes the normalized fields used by EvoMem memory
cards. It is designed to keep each stored item both
actionable at mutation time and auditable after a run. Not every field is
required for every card type: abstract idea cards emphasize reusable tactics
and explanations, while program cards additionally store concrete code
artifacts and fitness values.

\begin{center}
\small
\textbf{Memory-card schema.}\\[0.5em]
\begin{tabular}{p{0.26\linewidth}p{0.16\linewidth}p{0.45\linewidth}}
\toprule
\textbf{Field} & \textbf{Typical type} & \textbf{Purpose} \\
\midrule
\texttt{card\_id} & string & Stable identifier used for retrieval, merging, and usage tracking. \\
\texttt{card\_type} & enum & Distinguishes abstract idea cards from concrete program cards. \\
\texttt{description} & text & Concise statement of the reusable tactic or program-level finding. \\
\texttt{task\_summary} & text & Short description of the source task, constraints, and objective. \\
\texttt{metric\_context} & text & Description of the metric or fitness signal under which the card was observed. \\
\texttt{explanations} & list of text & Rationales, mechanism descriptions, and explanation summaries collected during analysis. \\
\texttt{keywords} & list of strings & Search terms used for exact and hybrid retrieval. \\
\texttt{source\_program\_ids} & list of strings & Program identifiers from which the idea or artifact was extracted. \\
\texttt{provenance} & structured record & Run identifier, generation, parentage, introduction events, and lineage evidence. \\
\texttt{aliases} & list of text & Earlier or alternative phrasings preserved during deduplication and merging. \\
\texttt{links} & list of strings & Optional links from an idea card to associated high-performing program cards. \\
\texttt{usage\_stats} & structured record & Counts, observed downstream fitness deltas, per-task summaries, and median delta. \\
\texttt{retrieval\_views} & structured text & Indexed views such as description, task summary, explanation summary, and their compositions. \\
\texttt{code} & optional text & Concrete evolved program body, present only for program cards. \\
\texttt{fitness} & optional number & Fitness value associated with a stored program card. \\
\bottomrule
\end{tabular}
\end{center}

\subsection{Write-phase pseudocode}
\label{app:write-pseudocode}

\paragraph{Algorithm 1: WriteMemory.}
\small
\begin{enumerate}[leftmargin=*]
    \item \textbf{Input:} completed run trace $R$, existing memory bank $M$.
    \item Collect completed programs from $R$ and remove root programs, invalid outputs, and entries without usable fitness or mutation metadata.
    \item Normalize each remaining program into a candidate record containing fitness, generation, parentage, task context, mutation descriptions, rationales, and code.
    \item Extract candidate ideas from the normalized records using either the default analyzer or the fast clustering analyzer.
    \item Enrich extracted ideas with task summaries, keywords, explanation summaries, aliases, and source-program identifiers.
    \item Estimate observational usage statistics for previously selected memories by comparing each child fitness with the strongest parent fitness.
    \item Run origin analysis to identify high-value ideas supported by introduction gains, sibling comparisons, descendant spread, and downside-rate checks.
    \item Normalize abstract idea cards and optional high-performing program cards into the memory-card schema.
    \item For each normalized card:
    \begin{enumerate}
        \item If its identifier already exists in $M$, update the existing entity and merge provenance and usage statistics.
        \item Else if the card is a concrete program, insert it directly.
        \item Else retrieve near-neighbor idea cards using multiple textual views and combine candidate scores.
        \item Ask the deduplication policy to store, discard, or merge the incoming idea.
        \item If the policy fails, store the incoming idea as a new card.
    \end{enumerate}
    \item Refresh retrieval indexes and persist the updated memory bank.
    \item \textbf{Output:} updated memory bank $M$ with write statistics and refreshed indexes.
\end{enumerate}
\normalsize

\subsection{Read-phase pseudocode}
\label{app:read-pseudocode}

\paragraph{Algorithm 2: RetrieveMemory.}
\small
\begin{enumerate}[leftmargin=*]
    \item \textbf{Input:} parent program $p$, task description $\tau$, metric description $\mu$, mutation context $x$, memory bank $M$.
    \item Validate the current program state and construct a structured memory request from $(p,\tau,\mu,x)$.
    \item If memory is disabled, return an empty memory section and an empty identifier list.
    \item Initialize the bounded selector loop with the request, the available memory abstracts, and the maximum number of retrieval iterations.
    \item While the selector has remaining iterations:
    \begin{enumerate}
        \item Plan one or more retrieval actions using exact keyword search, direct identifier lookup, or field-aware vector search.
        \item Execute each retrieval action with the configured top-$k$ limit.
        \item Merge hits by card identifier and keep the strongest evidence for each card.
        \item Reflect on the collected evidence and either stop with a final selection or issue a focused follow-up query.
    \end{enumerate}
    \item Convert the selected cards into concise mutation advice and truncate the resulting text to the prompt budget.
    \item Attach the selected card identifiers to the parent program metadata for later usage tracking.
    \item Insert the memory text into the mutation prompt as advice, leaving generation, validation, and scoring unchanged.
    \item \textbf{Output:} mutation prompt with a bounded memory section and selected memory identifiers.
\end{enumerate}
\normalsize

\section{Benchmarks Absolute Values}
\label{app:benchmark-absolute-values}
For completeness, Table~\ref{tab:absolute-values} reports the mean final target-metric values, on their original task-specific scales, across completed matched baseline--memory run pairs for each benchmark family summarized in the main text. Relative gains reported in Table~\ref{tab:results-summary} are calculated separately for each matched baseline--memory run pair and then averaged. Consequently, the mean relative gain need not equal the relative gain computed directly from the two mean values reported here.

\begin{table}[!htp]\centering\small
\caption{Absolute target-metric values for baseline and memory-enabled runs across benchmark families. Higher values indicate better performance for the reported task-specific target metric.}
\label{tab:absolute-values}
\resizebox{0.85\linewidth}{!}{%
\begin{tabular}{lrrr}\toprule
\textbf{Benchmark} &\textbf{Target metric baseline} &\textbf{Target metric with memory} \\\midrule
Circle Packing (32) &2.7870 &2.9289 \\
Circle Packing (26) &2.4218 &2.5577 \\
Heilbronn &0.0304 &0.0315 \\
Kissing Number (12D) &840.0000 &840.0000 \\
Hexagon Packing &-4.3424 &-4.1799 \\
HoVer &0.7900 &0.8080 \\
AlgoTune &461.5126 &495.3686 \\
KernelBench &2.4941 &2.7049 \\
HotpotQA &0.5389 &0.6015 \\
\bottomrule
\end{tabular}}
\end{table}

\section{Evolution tree and ideas visualization}
\begin{figure}[H]
    \centering
    \includegraphics[width=\textwidth]{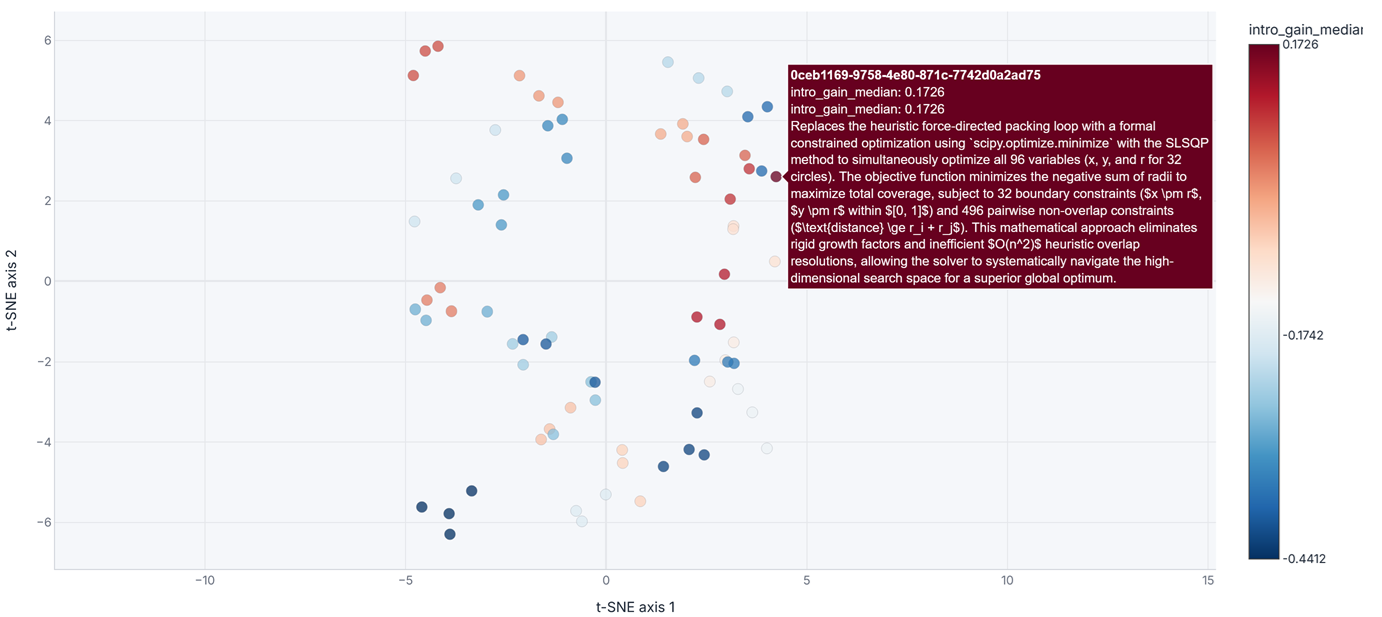}
    \vspace{0.5em}
    \includegraphics[width=\textwidth]{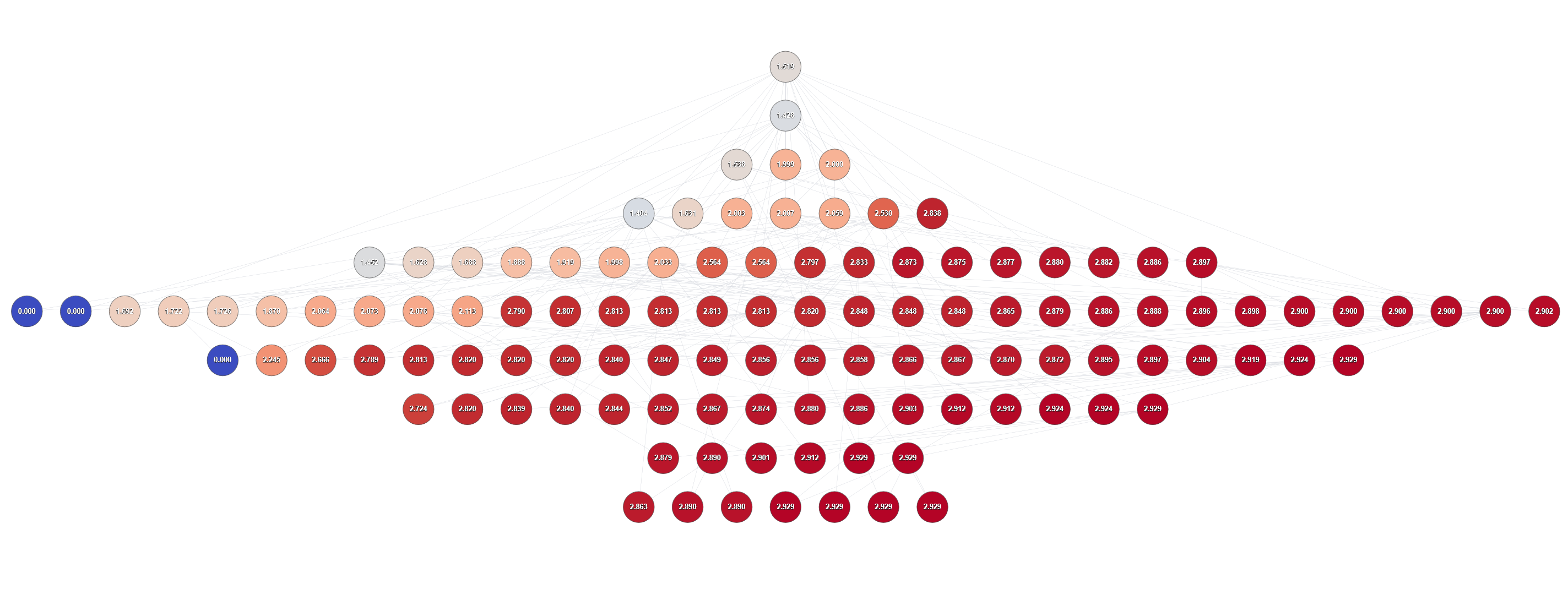}
    \caption{Top: visualization of extracted ideas on a two-dimensional plane obtained by applying t-SNE to embeddings of their textual descriptions. Bottom: evolution tree showing how programs branch and improve over the course of search.}
    \label{fig:results-visualizer}
\end{figure}

\section{Experimental Parameters}

\begin{itemize}
    \raggedright
    \item \textbf{Population size:} MAP-Elites island archive maximum size 75; initial population from all \path{*.py} files in each problem's \path{initial_programs/} directory; 5 elites per generation; 8 mutants per generation / epoch; at most 8 steady-state in-flight mutants.
    \item \textbf{Evaluation budget:} no fixed total evaluation budget by default; \path{max_generations} is null / unlimited by default, with most experiments using 30; per-generation mutant budget is 8, so the approximate mutant budget is \path{max_generations} \(\times 8\) plus initial and archive-refresh evaluations; \path{optimization_time_budget} is \(0.75 \times\) \path{dag_timeout}.
    \item \textbf{Memory top-\(k\):} \path{MemoryProvider} uses \path{max_cards=3}; memory runtime \path{search_limit=5}; Memory \path{Research Agent} allowed tools are \path{page_index} and \path{vector}; Memory \path{top_k_by_tool} uses \path{keyword=5}, \path{vector=3}, \path{vector_description=3}, \path{vector_task_description} \(\ge 1\), \path{vector_explanation_summary=3}, \path{vector_description_explanation_summary=3}, \path{vector_description_task_description_summary=3}, and \path{page_index=5}; card update deduplication retrieval uses \path{top_k_per_query=10}, \path{final_top_n=10}, and \path{min_final_score=0.05}.
    \item \textbf{Retrieval iterations:} Memory\path{ResearchAgent} uses \path{max_iters=3}.
    \item \textbf{Embeddings and clustering:} \path{IdeaTracker} uses \path{all-MiniLM-L6-v2}; memory uses embeddings-model \path{all-MiniLM-L6-v2}; DBSCAN uses \path{min_samples_for_dbscan=4}, \path{eps=0.25}, and \path{min_samples=2}; the fast analyzer uses \path{batch_size=32}, \path{max_attempts=20}, \path{max_rounds=100}, \path{recompute_center=false}, \path{refine_subgroup_size=10000}, and \path{llm_max_concurrent=100}.
    \item \textbf{Models and API:} main mutation, insights, and lineage model is \path{Gemini 3 Flash}; ideas-tracker analyzer model is \path{Gemini 3 Flash}; API base URL is \path{https://openrouter.ai/api/v1}; LLM defaults are \path{temperature=0.7}, \path{max_tokens=81920}, \path{top_p=0.95}, and \path{top_k=20}; many chain/prompt evaluations use \path{Qwen/Qwen3-8B} through local OpenAI-compatible endpoints.
    \item \textbf{Hardware:} NVIDIA A100-SXM4-80GB, driver 560.35.03; Intel Xeon Platinum 8358, 128 logical CPUs; Linux 5.15.0-1044-nvidia.
    \item \textbf{Runtime:} \path{stage_timeout=2400} s default; \path{dag_timeout=7200} s default; \path{dag_concurrency=16}; \path{max_concurrent_dags=10}; LLM request timeout 3200 s for all phases; stage and DAG timeouts are 7200 s for phase 1 and 18000 s for phases 2--3; generation timeout is 10800 s for phase 1 and 14400 s for phases 2--3.
    \item \textbf{Dataset splits and subsets:} no single universal dataset split; prompt tasks default \path{load_context} \path{n_samples=300}; HotpotQA chain defaults to the first 300 training samples, full validation uses \path{n_samples=200}, and optional train/validation and pool/held-out-validation splits use \path{seed=42}; GSM8K chain defaults to the first 200 training samples; test helpers commonly default to \path{n_samples=3} unless overridden.
\end{itemize}

\end{document}